\documentclass{article}
\usepackage{ijcai26}
\usepackage{times}
\usepackage{soul}
\usepackage{url}
\usepackage[hidelinks]{hyperref}
\usepackage[utf8]{inputenc}
\usepackage[small]{caption}
\usepackage{graphicx}
\usepackage{amsmath}
\usepackage{booktabs}
\usepackage{array}
\usepackage[switch]{lineno}
\usepackage{svg}
\usepackage{booktabs}
\usepackage{makecell}

\newcommand{\score}[2]{#1$\pm$ #2}
\title{Pretraining Language Models with Subword Regularization: An Empirical Study of BPE Dropout in Low-Resource NLP}

\author{
Ruan Visser, Trienko Grobler, Marcel Dunaiski
\affiliations
Department of Computer Science, Stellenbosch University, Stellenbosch, South Africa\\
\emails
ruanvisser101@gmail.com,
tlgrobler@sun.ac.za,
marceldunaiski@sun.ac.za
}

\begin{document}

\maketitle
\begin{abstract}
Subword regularization methods such as BPE dropout are typically applied only during fine-tuning, while pretraining is usually done with deterministic tokenization. This creates a potential word segmentation mismatch between pretraining and fine-tuning. We investigate whether applying BPE dropout during pretraining improves downstream performance in low-resource NLP. We trained monolingual and bilingual BERT models on downsampled subsets of English, German, French, Spanish, Kiswahili, and isiXhosa, and evaluated them on XNLI, PAWS-X, PAN-X, and MasakhaNER 2.0. Across these tasks, the best results are typically obtained when stochastic tokenization is applied during both pretraining and fine-tuning, whereas applying BPE dropout only during fine-tuning can underperform when compared to deterministic tokenization in smaller-data settings. As the amount of fine-tuning data increases, this disadvantage diminishes. We further find that the benefits of pretraining-time BPE dropout are largest when either pretraining or fine-tuning data is scarce, which suggests that its regularization and data augmentation effects might be particularly useful in resource-constrained settings.

The benefits of BPE dropout have often been attributed to better compositional representations, especially for rare words. To examine this, we also measured morphological boundary alignment under BPE dropout and found only modest improvements in expected alignment, while better-aligned segmentations remained rare. This suggests that fine-tuning alone may provide limited exposure to such segmentations, whereas stochastic tokenization during pretraining exposes the model to these segmentations more consistently. We further show that selectively introducing morphologically aligned segmentations during fine-tuning can improve performance for models pretrained without BPE dropout, while yielding smaller gains for models already exposed to stochastic segmentation during pretraining. Overall, these findings suggest that exposure to better-aligned segmentations may contribute to the downstream benefits observed when applying BPE dropout during pretraining.
\end{abstract}

\section{Introduction}

Subword tokenization has become the de facto approach for encoding text into discrete units usable by neural language models. Despite its widespread adoption, standard subword tokenization methods are typically deterministic, limiting representational diversity that may be beneficial in data-constrained settings. While numerous alternatives to conventional subword tokenizers have been proposed, many introduce substantial computational overhead while offering only modest downstream improvements~\cite{kreutzer_sokolov_2018,he_2020,meyer_2023}. For example, character- and byte-level models avoid fixed vocabularies but incur significantly longer sequence lengths, resulting in increased computational cost~\cite{lee_2017,xue_2022_byt5}. Neural tokenizers that jointly learn segmentation and language modeling can mitigate some of these issues without increasing sequence lengths, but often result in substantially slower pretraining and increased overall training complexity~\cite{he_2020,meyer_2023}.

Subword regularization introduces stochasticity into subword tokenization by exposing models to multiple valid segmentations of the same input during training~\cite{kudo_2018,provilkov_2020,wang_2021}. One commonly used variant of this approach is BPE dropout~\cite{provilkov_2020}, which stochastically drops merge operations during tokenization to produce alternative segmentations. By increasing representational diversity without modifying the underlying text, subword regularization occupies a middle ground between regularization and data augmentation, introducing bounded variability in the representation space. This bounded variability may be especially relevant when pretraining low-resource languages, which often comprise noisier training data that are orders of magnitude smaller than their high-resource counterparts.

However, existing work has typically applied subword regularization only during fine-tuning. It is therefore unclear what benefits stochastic tokenization during pretraining might provide, especially for low-resource languages. It may also introduce a pretraining--fine-tuning mismatch, as models are pretrained with deterministic tokenization but are exposed to stochastic segmentations only at the fine-tuning stage. As a result, the role of stochastic tokenization during pretraining remains underexplored. In particular, it is unclear whether applying subword regularization consistently across both pretraining and fine-tuning improves representation learning.

In this paper, we analyze the impact of applying BPE dropout during pretraining and fine-tuning, which we denote as PTD and FTD, in encoder-based language models under resource-constrained settings. We evaluate monolingual and bilingual models across English, Spanish, French, German, Kiswahili, and isiXhosa on XNLI, PAWS-X, and NER (PAN-X and MasakhaNER 2.0). In addition to evaluating the downstream effects of PTD and FTD, we also analyze morphological boundary alignment under BPE dropout in order to probe a possible mechanism behind observed gains.

Our experiments show that, in data-constrained settings, applying BPE dropout during both pretraining and fine-tuning typically yields better results than applying it only during fine-tuning (FTD) or using deterministic tokenization throughout. These comparisons suggest that inconsistent use of BPE dropout across training stages can reduce performance, consistent with a pretraining--fine-tuning mismatch. We further find that the benefits of using BPE dropout during both phases are larger when either pretraining or fine-tuning data is limited.

To better understand a possible mechanism behind these gains, we present a morphology-based analysis of BPE dropout segmentations. One common explanation for the benefits of BPE dropout is that alternative segmentations can yield better compositional representations, particularly for rare words~\cite{provilkov_2020,kudo_2018}. We find only a modest increase in expected alignment, while better-aligned segmentations remain relatively rare. This suggests that such segmentations may be sampled too infrequently during fine-tuning alone for models without PTD to benefit from them consistently.

To probe this hypothesis more directly, we conduct an additional experiment in which we fine-tune models trained with and without PTD using morphologically aligned segmentations, thereby exposing the model to them more consistently during fine-tuning. On average, models pretrained without BPE dropout benefit more from this intervention than models pretrained with it. While limited, these results are consistent with the hypothesis that repeated exposure to informative alternative segmentations contributes to the gains observed when BPE dropout is applied during pretraining.

\section{Background}
\subsection{Data Augmentation and Regularization}
Introducing controlled stochasticity during training has long been used to improve model generalization, particularly when supervision is limited. In computer vision, simple transformations such as rotations, cropping, and other image manipulations are particularly effective, as they preserve most of the semantic content while introducing useful variability \cite{lecun_1998,simard_2003,shorten_2019}. In natural language processing (NLP), however, augmentation must be applied more carefully in order to maintain syntactic structure and semantic meaning.

A wide range of data augmentation methods have been proposed for NLP. Surface-level perturbations, including token deletion, synonym replacement, and character-level noise injection, are simple to apply and computationally inexpensive \cite{xie_2017,wei_2019,feng_2021}.  However, such methods explicitly modify the surface form of text and may introduce syntactic or semantic distortions.

More semantic-preserving approaches, such as back-translation and paraphrase generation, have been shown to yield empirical improvements, particularly in low-resource settings. These methods, however, incur substantial computational cost when applied to large-scale corpora and rely on high-quality translation or generation systems, which are often unavailable for low-resource languages and may introduce systematic biases \cite{sennrich_2016,edunov2018,feng_2021}.

In contrast to surface-level and semantic-preserving augmentation, tokenization-level augmentation methods introduce variability directly at the representation level, which exposes models to multiple alternative segmentations of the same input without altering its semantic content or incurring the computational cost of text generation. These approaches can act as both regularizer and a form of data augmentation~\cite{cognetta_2024}, as they preserve the input string while altering the model's effective representation. Some neural tokenization approaches attempt to address limitations of fixed, discrete tokenization by learning token representations or compositions jointly with the model. While intuitively appealing, jointly learning tokenization and model parameters substantially increases training complexity and wall-clock time~\cite{meyer_2023,he_2020,song_2024}.  Reported performance gains over subword model baselines are often modest, suggesting that some of these improvements might be attainable by training standard subword-based models for longer under a similar compute budget.

\subsection{Subword Regularization}

Subword regularization provides a lightweight mechanism for introducing stochasticity within a fixed vocabulary. Rather than applying a single deterministic tokenization, models are exposed to multiple valid subword segmentations of the same input during training, reducing overfitting to specific segmentation decisions. Originally proposed for neural machine translation~\cite{kudo_2018,provilkov_2020}, subword regularization has since been shown to benefit downstream classification tasks. Wang {\em et al.}~\cite{wang_2021}, for example, investigate subword regularization in multilingual models such as mBERT and XLM-R, demonstrating that incorporating stochastic segmentation, together with their Multi-view Subword Regularization (MVR) approach, yields consistent improvements of up to 2.5 points over deterministic tokenization on the XTREME benchmark~\cite{hu_2020}.

Prior work suggests that subword regularization can smooth brittle tokenization decisions which improves robustness for rare and poorly represented words~\cite{kudo_2018,provilkov_2020}. This is especially relevant for low-resource languages, which are often over-segmented by shared subword vocabularies in multilingual models. Additionally, stochastic segmentation produces many alternative splits, which can include splits that better align with underlying morpheme boundaries, thereby exposing the model to linguistically meaningful subword groupings without requiring explicit morphological annotation. This variability could also reduce reliance on any single, potentially suboptimal segmentation and encourage more robust representations.

\subsection{Cross-lingual Synergy}

Shaham {\em et al.}~\cite{shaham_2023} characterize multilingual interactions as either synergistic, where joint training benefits performance, or interfering, where multilinguality degrades it. They show that multilingual training can give rise to both positive and negative cross-lingual interactions, depending on model and data conditions.

Meyer and Buys~\cite{meyer_2024} demonstrate that subword segmentation decisions play a significant role in shaping synergy and interference in multilingual neural machine translation. In particular, they show that subword regularization can enhance synergistic interactions in multilingual settings and yield substantial benefits for low-resource languages, highlighting the sensitivity of cross-lingual transfer to tokenization choices.

One potential contributor to cross-lingual synergy is lexical or subword overlap between languages, which provides shared anchor points for multilingual representations. Cognates, words that share similar form and meaning across languages, can facilitate alignment by encouraging similar representations for semantically related inputs. Lexical overlap may also introduce interference through false cognates, which share surface form but differ in meaning. However, Kallini {\em et al.}~\cite{kallini_2025} show that retaining all cognates in bilingual GPT models consistently yields better performances, and that removing false cognates alone has little effect. Tokenization choices, such as BPE dropout, can influence this balance by determining how words are decomposed into shared subword units, effectively creating partial cognates that anchor related lexical forms across languages.

\section{Methodology}
\subsection{Data Mixtures}
We trained both monolingual and bilingual BERT models using BPE tokenizers with and without BPE dropout applied during pretraining. Our experiments cover four high-resource European languages, namely, English, Spanish, French, and German, which are shared across all primary downstream evaluation datasets considered: XNLI, PAWS-X, and PAN-X. In addition, we included two low-resource languages, Kiswahili and isiXhosa\footnote{isiXhosa training subsets do not exist for XNLI or PAWS-X.}. We trained each monolingual model using 1\,GB of text from the specific language. For bilingual settings, 1\,GB of English was paired with 100\,MB of a target language which allows us to also evaluate cross-lingual transfer under consistent data mixture and tokenization conditions. 

\subsection{Models}
All models are based on the BERT encoder architecture and follow standard masked language model pretraining practices. We adopt the hyperparameter configuration introduced by Devlin {\em et al.}~\cite{devlin2019}, with modifications to accommodate the scale of our datasets and available compute.

To reduce training cost, we limited the maximum input length to 128 tokens. The resulting decrease in tokens per batch is partially offset by increasing the batch size to 512, which we found sufficient to maintain stable optimization. Model capacity was chosen to align with the amount of available training data. Specifically, because each of our corpora is approximately 1\,GB in size, we employed base-sized models in accordance with prior scaling analyses~\cite{visser2025}. Furthermore, following the observations made by Visser {\em et al.} that extended training can be beneficial in data-constrained settings, we trained the 1\,GB and bilingual models for one million update steps and the 100\,MB models for 500 thousand update steps. This is particularly important in our setting where stochastic variation, introduced through subword regularization, increases the effective diversity of the training signal.

\subsection{Tokenizer}
We trained uncased BPE tokenizers with a vocabulary size of 30,000 for each model, matching BERT's vocabulary size~\cite{devlin2019}. For our bilingual models we trained the BPE tokenizers on 100MB of both English and the target language to ensure a balanced vocabulary. We then added the remaining 900\,MB of English before training each language model. We experimented with various BPE dropout probabilities for both pretraining and fine-tuning and found that a probability of 5\% performed best for pretraining and 10\% for fine-tuning, and consequently used these probabilities for all our experiments. 

\subsection{Morphological Alignment}
\label{sec:meth_gold_boundary}

\subsubsection{Evaluation}
One potential benefit of BPE dropout is that it exposes the model to segmentations that may better align with morphological boundaries~\cite{provilkov_2020}. Arnett and Bergen~\cite{arnett_bergen_2025} proposed MorphScore as a measure of morphological alignment. MorphScore uses annotated boundaries for a word and assigns a score of 1 when a gold boundary is recovered and 0 otherwise. However, it does not penalize incorrect boundaries. Since such extra boundaries are likely to arise under BPE dropout, we instead use the boundary-level F1 metric proposed by Poelman {\em et al.}~\cite{poelman_2025}, which more evenly balances missed and incorrect boundaries.

For morphological boundaries, we use MorphyNet~\cite{batsuren_2021_morphynet} entries for English, French, German, and Spanish, and match these words against the words occurring in the PAWS-X training set. Because a word can have multiple valid morphological segmentations, for example reflecting inflectional or derivational structure, we construct a set of valid gold boundary patterns for each word and score each segmentation against the best-matching gold pattern, i.e., the one that yields the highest F1 score.

More formally, for each word \(w\), let \(G(w)\) be the set of valid gold boundary sets. For any segmentation \(\hat{s}\), let \(B(\hat{s})\) denote its boundary set. We then define the scoring function as
\[
\phi(w,\hat{s})=\max_{g\in G(w)} F1\!\left(B(\hat{s}),g\right).
\]

Deterministic alignment is computed from the regular BPE segmentation \(\hat{s}^{\mathrm{reg}}_w\):
\[
F1_{\mathrm{reg}}(w)=\phi\!\left(w,\hat{s}^{\mathrm{reg}}_w\right).
\]

For BPE dropout, let \(\{\hat{s}^{(i)}_w\}_{i=1}^{N}\) be \(N\) sampled segmentations (here \(N=2000\)). The expected dropout alignment is
\[
F1_{\mathrm{drop}}(w)=\frac{1}{N}\sum_{i=1}^{N}\phi\!\left(w,\hat{s}^{(i)}_w\right).
\]
To assess whether better-aligned segmentations are available within the BPE dropout distribution, we also report constrained oracle scores. For each word, \texttt{Best$\geq$100} and \texttt{Best$\geq$10} denote the highest boundary-level F1 among sampled segmentations observed at least 100 or 10 times, respectively.

\subsubsection{Aligned Fine-tuning}
Other than token-pair frequency, BPE dropout has no explicit mechanism for producing morphologically or linguistically meaningful units~\cite{bauwens_delobelle_2024_bpe}. While BPE dropout during pretraining can expose the model to segmentations that better align with morphological boundaries, a fine-tuning-only alternative that increases exposure to such segmentations would be preferable.

To test whether more consistent exposure to morphologically aligned segmentations can reproduce some of the gains associated with PTD, we inject MorphyNet-derived split replacements during fine-tuning. 
A word is eligible for replacement only if it has a fully aligned segmentation compatible with the tokenizer vocabulary, where each segment maps cleanly to a single tokenizer unit and the resulting segmentation decodes back to the original surface form. During fine-tuning, each eligible word is replaced with one randomly selected valid segmentation with a probability of 25\%\footnote{We evaluated four different replacement probabilities 10\%, 25\%, 50\%, 100\% and found that a replacement probability of 25\% performed best.}.

\subsection{Fine-tuning Tasks}
We evaluated pretrained models on three downstream tasks. For sentence-level understanding, we used XNLI~\cite{conneau2018xnli} and PAWS-X~\cite{pawsx}, while token-level performance was assessed using the PAN-X~\cite{xtreme} and MasakhaNER 2.0~\cite{adelani_2022_masakhaner20} named entity recognition benchmarks.

XNLI consists of translated training and evaluation sets derived from the English MNLI~\cite{williams_2018_mnli} (Multi-Genre Natural Language Inference) dataset\footnote{https://huggingface.co/datasets/facebook/xnli}. The task requires models to classify the relationship between a premise and a hypothesis as entailment, contradiction, or neutral. MNLI is a large dataset containing approximately 393{,}000 sentence pairs spanning ten distinct domains, ensuring genre diversity during training. Due to its size, most experiments use a version of XNLI downsampled to 10\% of the original training data to better approximate lower-resource fine-tuning conditions. We additionally evaluate on the full dataset when analyzing the interaction between BPE dropout and training data size.

Similar to XNLI, PAWS-X is a sentence-level task based on the English PAWS (Paraphrase Adversaries from Word Scrambling) dataset~\cite{zhang_2019_paws}. Instead of predicting semantic relations, models must determine whether two sentences are paraphrases. PAWS-X is smaller than XNLI, with approximately 49{,}000 training instances per language. However, many examples are slight variations of the same underlying sentence pairs, and the paired sentences are intentionally highly similar by design, which means that the effective diversity of the training set is substantially lower than the raw instance count suggests. We also experiment with fine-tuning on 10\% of PAWS-X to better simulate lower-resource settings.

For sequence labelling, we used the PAN-X and MasakhaNER 2.0 named entity recognition datasets. PAN-X is derived from the WikiANN corpus and provides automatically projected BIO-style annotations across many languages, with approximately 20{,}000 training instances per language. MasakhaNER 2.0 is a human-annotated benchmark for African languages and contains approximately 6{,}600 training instances for Kiswahili and 5{,}700 for isiXhosa.

For each task, we report results averaged over 10 independent fine-tuning runs with different random seeds. We largely followed standard fine-tuning settings, with task-specific adjustments where necessary. We performed a grid search over learning rates $\{2\times10^{-5}, 5\times10^{-5}, 1\times10^{-4}\}$ and selected $2\times10^{-5}$ based on validation performance.

Furthermore, we tuned training duration and learning-rate warmup as part of hyperparameter selection. Based on validation performance, we selected task-specific fine-tuning schedules of 4 epochs for NER, 3 epochs for XNLI, and 5 epochs for PAWS-X. For downsampled XNLI, we trained for 6 epochs and did not observe meaningful gains from further increasing the number of epochs. Similarly, for downsampled PAWS-X, we doubled the number of epochs used for the full-dataset and trained for 10 epochs. To improve optimization stability across runs, we applied a weight decay of 0.01 and a warmup ratio of 0.1 in all experiments.

\begin{figure*}[t]
  \centering
  \includegraphics[width=\textwidth]{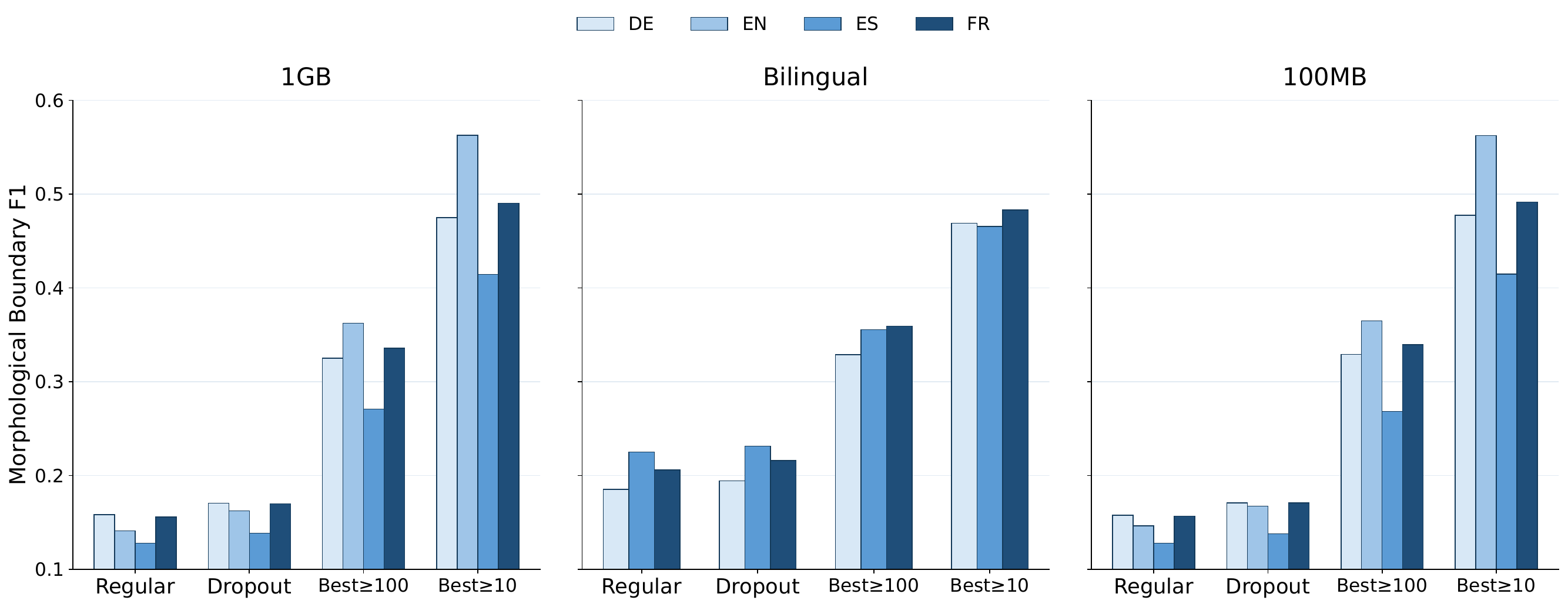}
  \caption{Morphological boundary F1 comparison for deterministic tokenization (Regular), expected BPE dropout (Dropout), and the best observed sampled segmentation among candidates occurring at least 100 or 10 times (\texttt{Best$\geq$100} and \texttt{Best$\geq$10}).}
  \label{fig:morph-boundary-f1}
\end{figure*}

\section{Results}

\subsection{Linguistic Alignment of BPE vs BPE Dropout}

\begin{table}[t]
  \centering
  \begin{tabular}{lrr}
    \hline
    Language & 1GB \% & Bilingual \% \\
    \hline
    German & 63.45 & 75.23 \\
    Spanish & 59.03 & 72.98 \\
    French & 59.62 & 70.83 \\
    \hline
  \end{tabular}
\caption{Percentage of words in the MorphyNet--PAWS-X overlap that are split into multiple subword units by the 1GB monolingual and bilingual tokenizers.}
  \label{tab:multi-token-increase}
\end{table}

We analyze the effect of BPE dropout on morphological boundary alignment using the boundary-level F1 metric described in Section~\ref{sec:meth_gold_boundary}. Figure~\ref{fig:morph-boundary-f1} reports the expected boundary-level F1 before (\texttt{Regular}) and after applying BPE dropout (\texttt{Dropout}), along with the best observed F1 for each word among sampled segmentations occurring at least 100 and 10 times, labeled \texttt{Best$\geq$100} and \texttt{Best$\geq$10}, respectively.

For deterministic tokenization, we observe similar F1 scores for tokenizers trained on 1GB and 100MB of data, suggesting that, in our setting, morphological alignment of tokenizers are relatively stable for these training data volumes when the data is drawn from the same distribution.

In contrast, bilingual tokenizers consistently yield higher alignment scores than monolingual models. One likely contributor is that sharing a fixed vocabulary across languages forces more words to be segmented into multiple subword units, increasing the chance of recovering morphological boundaries that would otherwise remain within single tokens. Table~\ref{tab:multi-token-increase} is consistent with this explanation, as bilingual tokenizers increase the proportion of multi-token words relative to the 1GB monolingual tokenizer. However, this factor alone does not appear to fully explain the observed gains. The increase in multi-token words is modest (18--24\%), whereas the alignment improvements vary substantially across languages. In particular, Spanish shows a much larger increase in alignment (+75.97\%) than German (+17.02\%) and French (+32.29\%).

Figure~\ref{fig:morph-boundary-f1} shows a modest increase in expected morphological alignment under BPE dropout. This is broadly consistent with Bauwens and Delobelle~\cite{bauwens_delobelle_2024_bpe}, who find that BPE dropout does not necessarily improve overall morphological adherence. At the same time, the \texttt{Best$\geq$10} results indicate that substantially better-aligned segmentations are sampled, but occur less frequently. This suggests that many of the most morphologically aligned segmentations may be sampled only rarely during fine-tuning, especially for lower-frequency words. As a result, they may provide too little signal for the model to benefit from them reliably during fine-tuning. BPE dropout during pretraining could mitigate this by exposing the model to low-probability segmentations that would otherwise remain relatively novel at fine-tuning time. Alternatively, if morphological segmentations are known beforehand and can be introduced more often during fine-tuning, they may also reduce the novelty of morphologically aligned words at the fine-tuning stage.

In the next sections, we examine the downstream benefits of BPE dropout for low-resource language models, the interaction between PTD and FTD, and the impact of morphologically aligned segmentations derived from MorphyNet on downstream performance.

\begin{figure}[t]
  \centering
  \includegraphics[width=\linewidth]{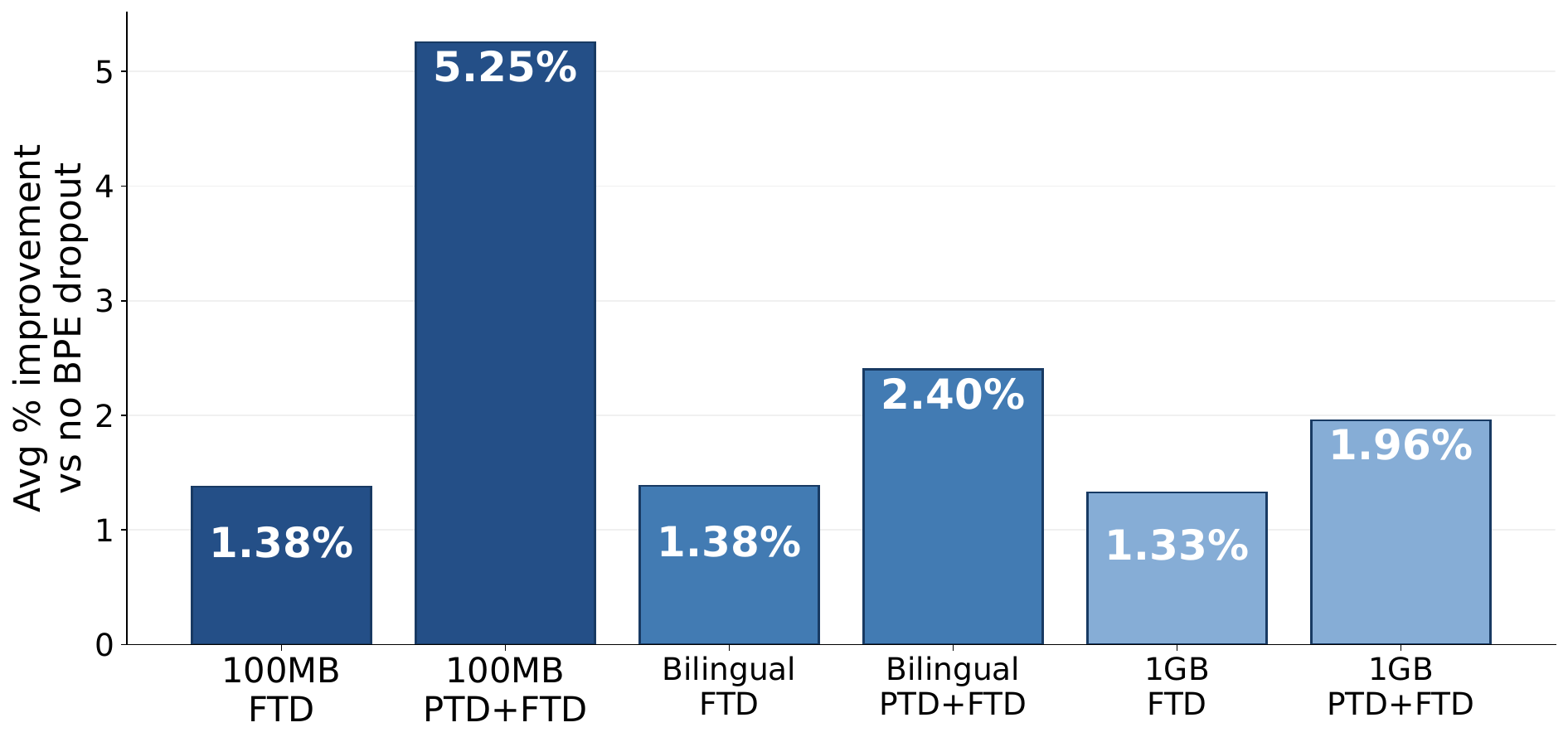}
  \caption{Average relative improvement on downsampled XNLI across languages over the baseline with deterministic tokenization during both pretraining and fine-tuning (no BPE dropout), for models fine-tuned with BPE dropout (FTD), shown for 100MB monolingual, bilingual, and 1GB monolingual settings pretrained with and without BPE dropout (PTD+FTD or FTD).}
  \label{fig:results-pulled-xnli-all}
\end{figure}

\begin{figure}[t]
  \centering
  \includegraphics[width=\linewidth]{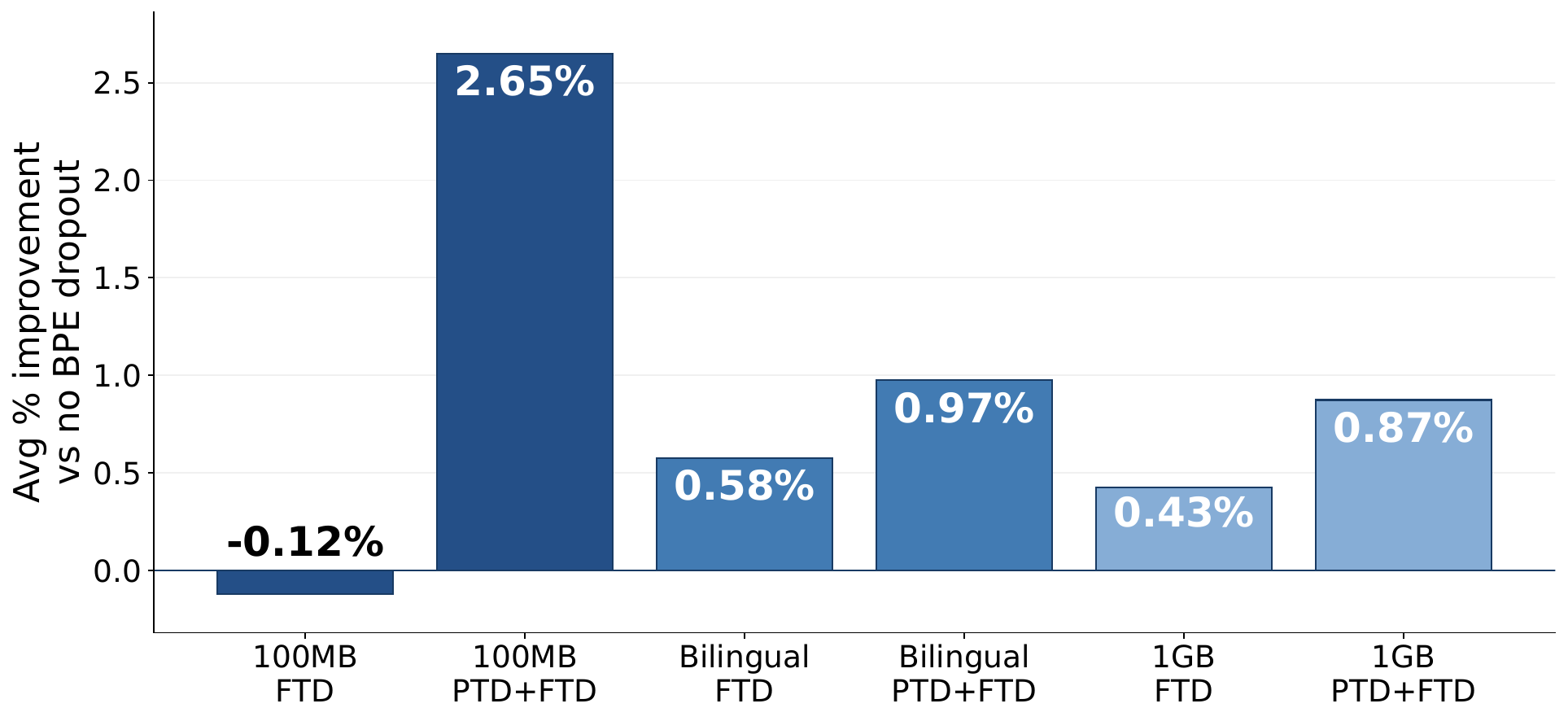}
  \caption{Average relative improvement on PAWS-X across languages over the baseline with deterministic tokenization during both pretraining and fine-tuning (no BPE dropout), for models fine-tuned with BPE dropout (FTD), shown for 100MB monolingual, bilingual, and 1GB monolingual settings pretrained with and without BPE dropout (PTD+FTD or FTD).}
  \label{fig:results-pulled-pawsx-all}
\end{figure}

\begin{figure*}[t]
  \centering
  \includegraphics[width=0.65\linewidth]{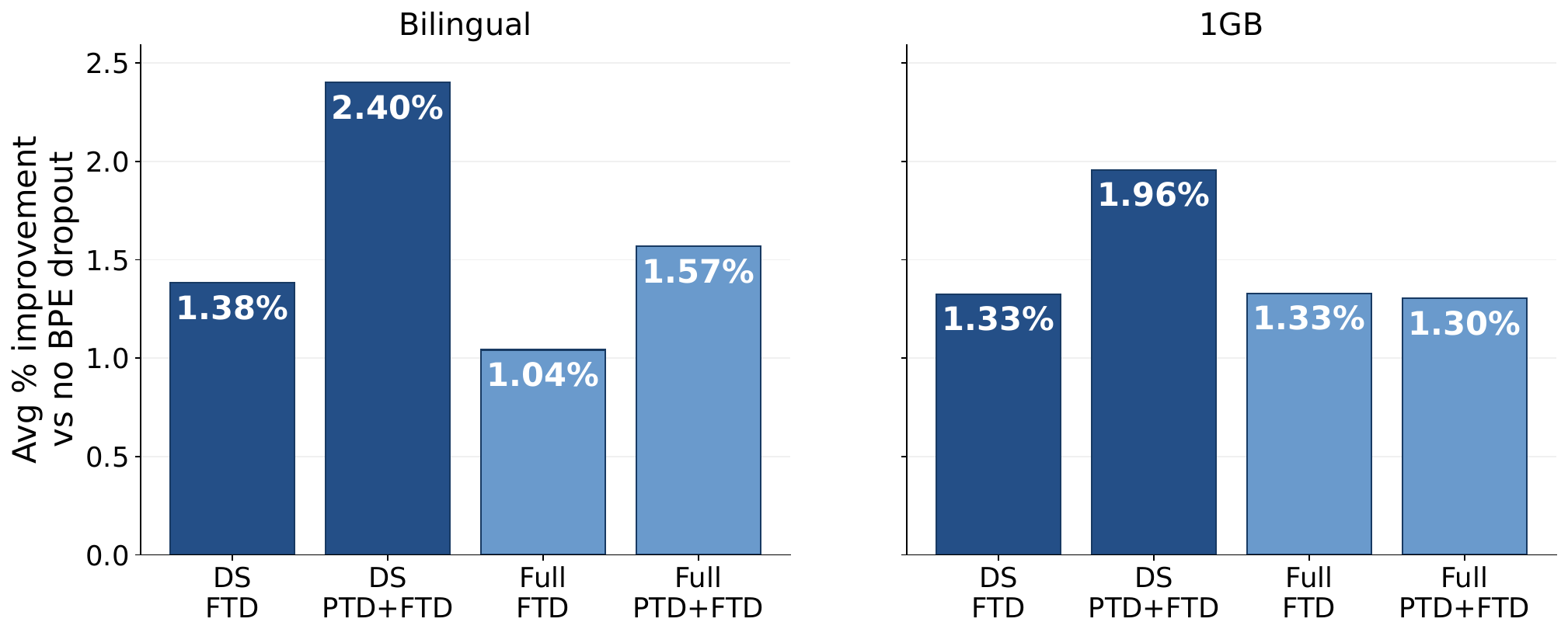}
  \caption{Average relative improvement on XNLI across languages over the baseline with deterministic tokenization during both pretraining and fine-tuning (no BPE dropout), comparing downsampled (DS) and full fine-tuning data (Full) in the bilingual and 1GB monolingual settings.}
  \label{fig:results-pulled-xnli-two}
\end{figure*}

\begin{figure}[t]
  \centering
  \includegraphics[width=\linewidth]{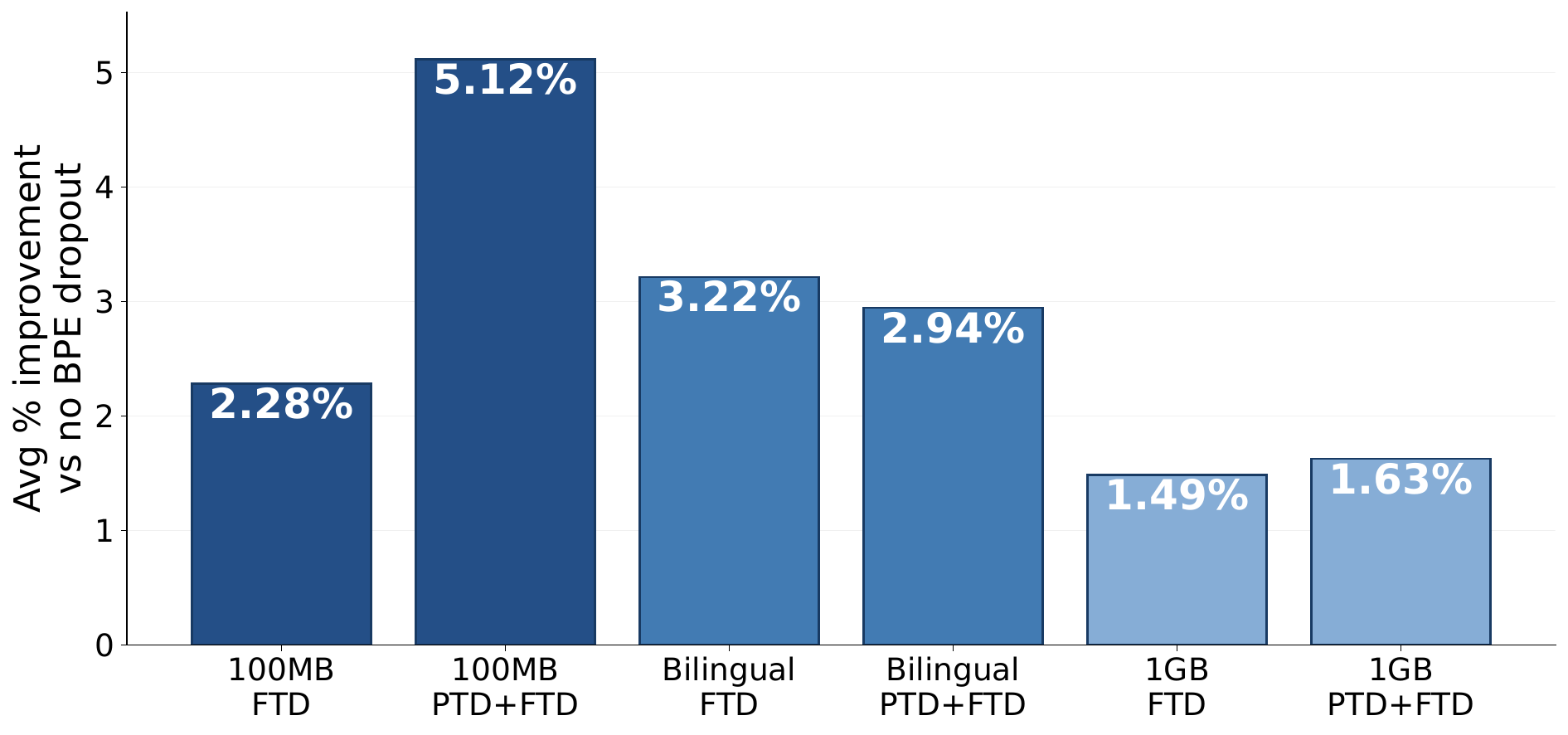}
  \caption{Average relative improvement on downsampled Kiswahili XNLI over the deterministic-tokenization baseline (no BPE dropout), shown for 100MB monolingual, bilingual, and 1GB monolingual settings under FTD, with and without PTD.}
  \label{fig:results-swahili-xnli-downsample}
\end{figure}

\begin{figure}[t]
  \centering
  \includegraphics[width=1.03\linewidth]{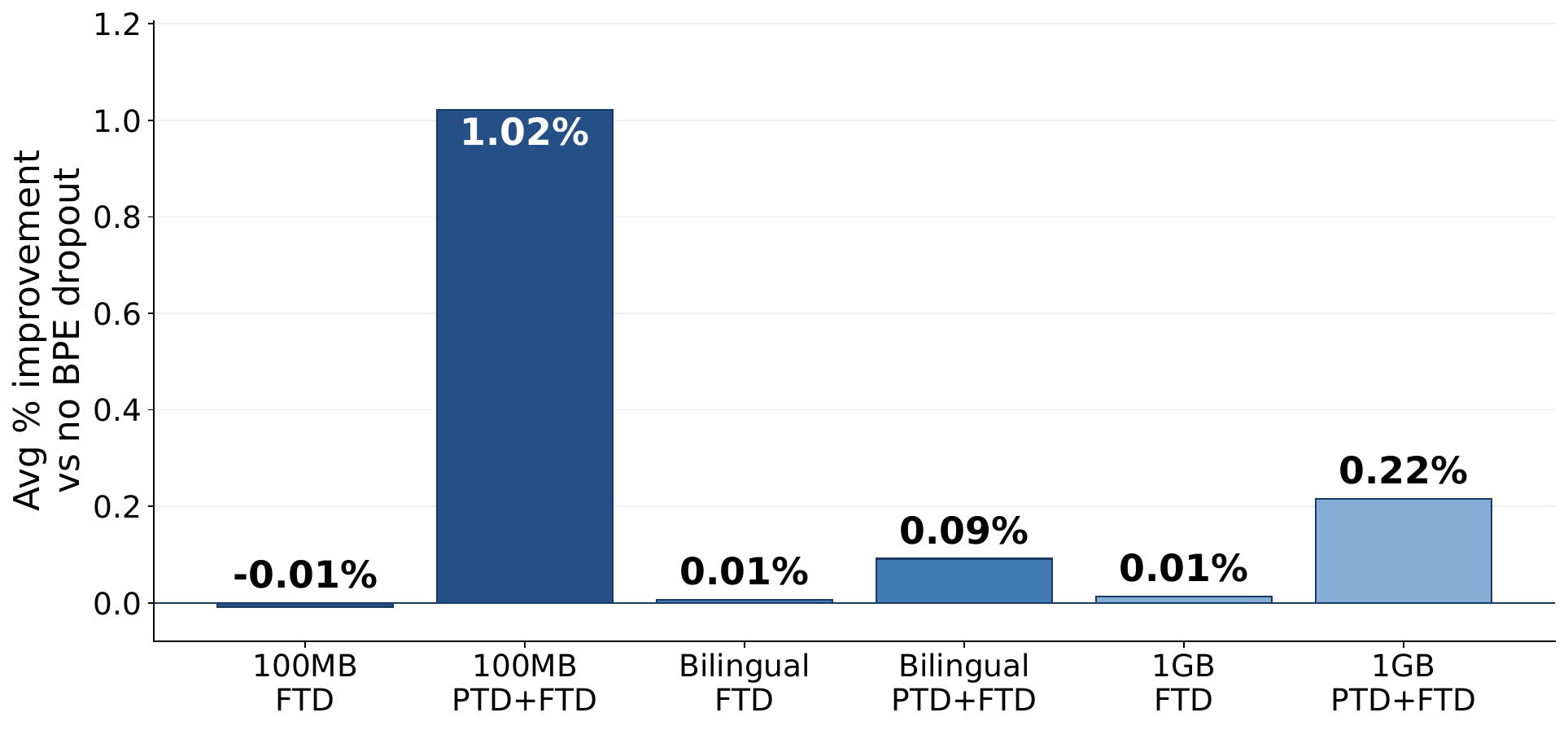}
  \caption{Average relative improvement in XTREME and MasakhaNER2.0 performance over the baseline with deterministic tokenization during both pretraining and fine-tuning (no BPE dropout), for 100MB monolingual, bilingual, and 1GB monolingual settings under FTD, with and without PTD.}
  \label{fig:results-ner}
\end{figure}

\subsection{PTD and FTD} 
\label{sec:ptd_ftd}

Applying BPE dropout during both pretraining and fine-tuning (PTD+FTD) yields the strongest average downstream performance overall, whereas fine-tuning-only dropout (FTD) produces smaller and less consistent gains (see Figures~\ref{fig:results-pulled-xnli-all} and~\ref{fig:results-pulled-pawsx-all}). The advantage of PTD+FTD is greatest when pretraining data is limited, suggesting that stochastic segmentation is especially beneficial in resource-constrained settings. This pattern is evident in the averaged XNLI and PAWS-X results and for the low-resource language Kiswahili (see Figure~\ref{fig:results-swahili-xnli-downsample}), with the largest improvements in the 100\,MB monolingual setting and smaller gains in the bilingual and 1\,GB monolingual settings.

For NER, gains from fine-tuning-only dropout (FTD) are negligible across languages and pretraining data volumes, whereas PTD+FTD yields a clear improvement in the 100\,MB setting and smaller gains in the 1\,GB and bilingual settings for selected languages (see Figure~\ref{fig:results-ner}).

\subsection{Downsampled Performance} 

In Section~\ref{sec:ptd_ftd}, we showed that BPE dropout during pretraining is most beneficial under resource-constrained pretraining conditions, particularly in the 100\,MB setting. However, low-resource settings often also involve limited fine-tuning data. Figure~\ref{fig:results-pulled-xnli-two} shows that this trend extends to fine-tuning: PTD+FTD yields larger improvements when the amount of fine-tuning data is reduced, consistent with the pretraining comparisons in Figures~\ref{fig:results-pulled-xnli-all} and~\ref{fig:results-pulled-pawsx-all}.

\begin{figure*}[t]
  \centering
  \includegraphics[width=0.75\linewidth]{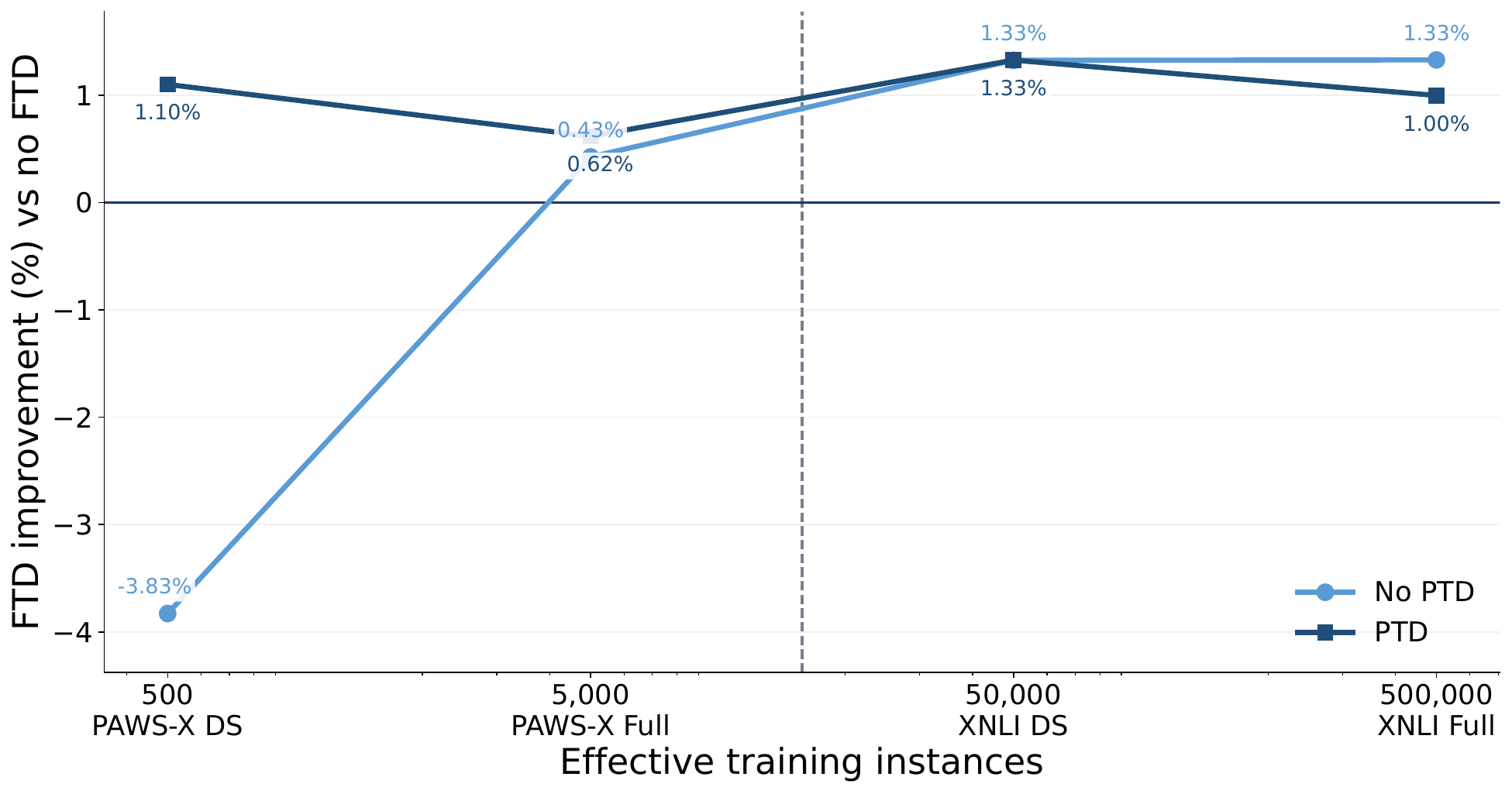}
  \caption{Incremental gain from adding fine-tuning-time BPE dropout (FTD) in the 1GB monolingual setting, plotted against the effective number of fine-tuning instances for PAWS-X and XNLI. Curves compare models pretrained with and without PTD.}
  \label{fig:results-ftd-1gb-full-ds}
\end{figure*}

Figure~\ref{fig:results-ftd-1gb-full-ds} isolates the gains from fine-tuning-only dropout in 1\,GB monolingual models pretrained with and without PTD. We compare four fine-tuning data scales, in increasing order: PAWS-X DS (500), PAWS-X (5\,000), XNLI DS (50\,000), and XNLI (500\,000)\footnote{Given PAWS-X's limited effective training diversity, we treat full PAWS-X as roughly comparable to a 5{,}000-example fine-tuning set when comparing data scale.}. For PTD-pretrained models, the gains from FTD are largest when the fine-tuning dataset is smallest and decrease as more fine-tuning data becomes available. For models pretrained without PTD, the gains from FTD become more similar and eventually exceed those of the PTD-pretrained models at larger fine-tuning data scales. A similar pattern is observed in the bilingual setting (Figure~\ref{fig:appendix-ftd-bilingual-full-ds}).

\section{Discussion}

\subsection{Pretraining–Fine-tuning Mismatch}

The benefit of subword segmentation and BPE dropout is often attributed to improved compositional learning, especially for rare words~\cite{provilkov_2020,kudo_2018,he_2020}. One possible mechanism is that some sampled segmentations align more closely with morphological boundaries, making the resulting subword units more linguistically meaningful (and robust building blocks). However, in Section~\ref{sec:meth_gold_boundary}, we find that many of the most morphologically aligned segmentations occur fewer than 100 times in 2000 samples, i.e., less than 5\% of the time. For rare words, especially in smaller fine-tuning sets, such segmentations may therefore be too infrequent for the model to benefit from them consistently. This may help explain why downsampled PAWS-X, our smallest fine-tuning dataset, is particularly sensitive to pretraining--fine-tuning mismatch: deterministically pretrained models that are fine-tuned with BPE dropout (FTD only) perform worse than models trained deterministically during both phases, whereas models exposed to BPE dropout during both pretraining and fine-tuning (PTD+FTD) perform best by a significant margin. By contrast, larger fine-tuning datasets can expose the model to more segmentation variants, which may reduce the need for PTD. 

\subsection{Regularization and Data Augmentation}
Subword regularization, a class of methods that includes BPE dropout, is commonly described as both a regularization strategy and a form of data augmentation~\cite{cognetta_2024,provilkov_2020,kudo_2018}.

We observe that models using both PTD and FTD are more effective in data-constrained settings, for both pretraining and fine-tuning, suggesting that stochastic tokenization can improve representation robustness and generalization when data is limited. This indicates that BPE dropout may be especially useful in lower-resource settings, where pretraining text and fine-tuning data are typically scarce.

Mroczkowski {\em et al.}~\cite{mroczkowski-2021} report limited and statistically insignificant gains from pretraining with BPE dropout. While their setup differs from ours in several respects, one possible explanation is that even their smaller corpus is relatively large (1,658M tokens compared to the roughly 220M tokens in our 1GB dataset). Because their experiments also use a base-sized model, the larger data volume may reduce the need for additional regularization.

\subsection{Morphologically aligned replacements}
\label{sec:result_gold_boundary_ft}

The alignment analysis above showed that BPE dropout can make better-aligned segmentations available, but that many of these segmentations occur only rarely (Figure~\ref{fig:morph-boundary-f1}). We therefore test whether some of the gains associated with PTD can be reproduced by explicitly exposing the model more consistently to a smaller set of morphology-aligned segmentations during fine-tuning.

Table~\ref{tab:pawsx-xnli-short-with-005} reports results for monolingual 1GB models on PAWS-X and downsampled XNLI. On average, MorphyNet replacements improve models pretrained without PTD by 0.403\% on PAWS-X and 0.576\% on XNLI, while the gains are smaller for models pretrained with PTD (0.129\% and 0.139\%, respectively). This suggests that targeted exposure to morphologically aligned segmentations can partially compensate for the absence of PTD, but provides less additional benefit when similar compositional variation has already been observed during pretraining.

Although these results are limited, they are consistent with the view that part of PTD's benefit arises from repeated exposure to informative alternative segmentations. At the same time, MorphyNet-based replacements require external morphological resources, apply only to a subset of words, and contain a limited number of languages. PTD is therefore the more general and practical approach, especially in lower-resource settings.

\begin{table}[t]
\centering
\small
\begin{tabular}{llcc}
\toprule
Language & PTD & PAWSX & XNLI \\
\midrule
de & 0.00 & +0.562\% & +0.423\%  \\
de & 0.05 & +0.392\% & +0.368\%  \\
en & 0.00 & -0.095\% & +0.378\%  \\
en & 0.05 & -0.319\% & -0.553\%  \\
es & 0.00 & +0.873\% & +0.675\%  \\
es & 0.05 & +0.171\% & +0.762\%  \\
fr & 0.00 & +0.274\% & +0.829\%  \\
fr & 0.05 & +0.273\% & -0.021\%   \\
\midrule
Avg gain & 0.00  & +0.403\% & +0.576\%  \\
Avg gain & 0.05  & +0.129\% & +0.139\%  \\
\bottomrule
\end{tabular}
\caption{Percentage improvement from MorphyNet-based fine-tuning replacements, relative to the corresponding no-replacement, no-FTD baseline within each PTD setting, shown by language for monolingual 1GB models on PAWS-X and downsampled XNLI under no PTD (0.00) and PTD (0.05).}
\label{tab:pawsx-xnli-short-with-005}
\end{table}

\section{Conclusion}

We investigated the role of BPE dropout during pretraining for encoder-based language models in resource-constrained settings. Across tasks and languages, the best results were typically obtained when stochastic tokenization was applied during both pretraining and fine-tuning, especially in lower-resource settings. Applying BPE dropout only during fine-tuning can have a small beneficial impact, however, in lower-resource settings it can underperform deterministic tokenization. These findings are consistent with a pretraining--fine-tuning mismatch and suggest that pretraining-time BPE dropout is particularly useful when either pretraining or fine-tuning data is limited.

Our analysis of morphological boundary alignment indicates that BPE dropout does not substantially improve expected alignment on average, although better-aligned segmentations do occur within its sampling distribution. Together with our additional fine-tuning intervention, these results suggest that repeated exposure to informative alternative segmentations may contribute to the benefits of pretraining-time BPE dropout.

Overall, our findings show that BPE dropout during pretraining is a simple, low-overhead way to improve robustness in low-resource settings, especially when stochastic tokenization is used consistently across training stages. We suggest that future work use the distributional evaluation of morphological alignment introduced in Section~\ref{sec:meth_gold_boundary} to compare different subword regularization methods and investigate whether stochastic segmentation strategies can be made more morphologically aligned without relying on external morphological resources.

\section{Acknowledgments}
This work was supported by Google’s TPU Research Cloud program and partially funded by MICT-SETA. The author also acknowledges an Apple travel grant, which supported the broader research activities underlying this work.

\bibliographystyle{named}
\bibliography{main}

\appendix

\section{Appendix}

\begin{figure*}[t]
  \centering
  \includegraphics[width=0.75\linewidth]{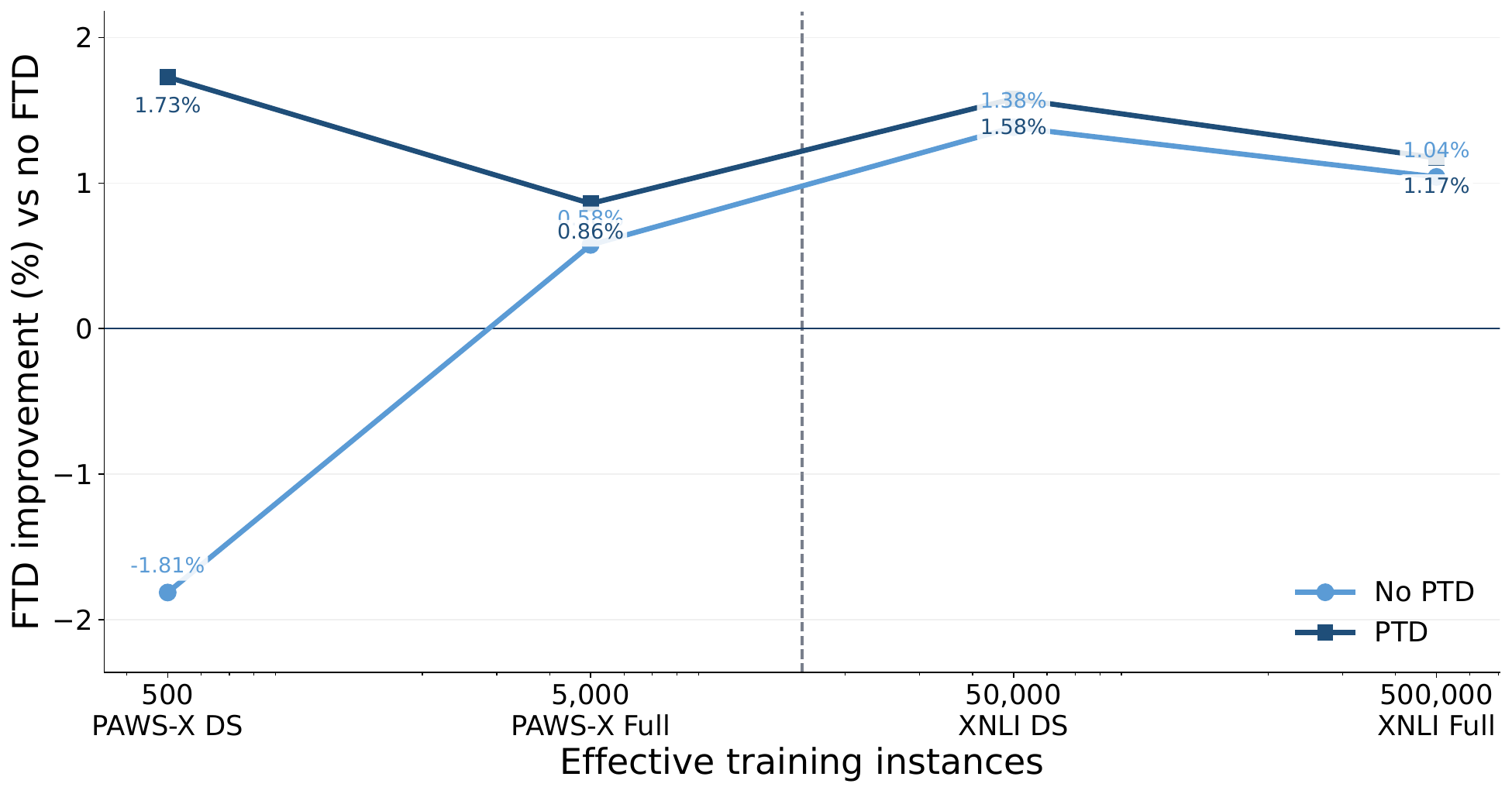}
  \caption{Incremental gain from adding fine-tuning-time BPE dropout (FTD) in the bilingual setting, plotted against the effective number of fine-tuning instances for PAWS-X and XNLI. Curves compare models pretrained with and without PTD.}
  \label{fig:appendix-ftd-bilingual-full-ds}
\end{figure*}
\begin{table*}[t]
\centering
\scriptsize
\setlength{\tabcolsep}{5pt}
\renewcommand{\arraystretch}{1.08}
\begin{tabular}{llcccc}
\toprule
& & \multicolumn{2}{c}{Deterministic pretraining} & \multicolumn{2}{c}{PTD pretraining} \\
\cmidrule(lr){3-4}\cmidrule(l){5-6}
Language & PT setup & No FTD & +FTD & No FTD & +FTD \\
\midrule
de & 100MB     & \score{0.5360}{0.0080} & \score{0.5431}{0.0065} & \score{0.5465}{0.0115} & \score{0.5639}{0.0142} \\
de & Bilingual & \score{0.6550}{0.0055} & \score{0.6664}{0.0043} & \score{0.6540}{0.0031} & \score{0.6676}{0.0039} \\
de & 1GB       & \score{0.6667}{0.0053} & \score{0.6779}{0.0056} & \score{0.6772}{0.0048} & \score{0.6838}{0.0045} \\
\addlinespace[2pt]
en & 100MB     & \score{0.6782}{0.0046} & \score{0.6853}{0.0042} & \score{0.6969}{0.0048} & \score{0.6998}{0.0040} \\
en & 1GB       & \score{0.7463}{0.0050} & \score{0.7528}{0.0031} & \score{0.7486}{0.0052} & \score{0.7564}{0.0042} \\
\addlinespace[2pt]
es & 100MB     & \score{0.5430}{0.0101} & \score{0.5416}{0.0148} & \score{0.5591}{0.0081} & \score{0.5682}{0.0092} \\
es & Bilingual & \score{0.6770}{0.0020} & \score{0.6846}{0.0037} & \score{0.6915}{0.0033} & \score{0.7001}{0.0060} \\
es & 1GB       & \score{0.6955}{0.0045} & \score{0.7029}{0.0037} & \score{0.6931}{0.0049} & \score{0.7030}{0.0061} \\
\addlinespace[2pt]
fr & 100MB     & \score{0.5381}{0.0081} & \score{0.5564}{0.0065} & \score{0.5453}{0.0089} & \score{0.5809}{0.0062} \\
fr & Bilingual & \score{0.6684}{0.0048} & \score{0.6771}{0.0025} & \score{0.6713}{0.0049} & \score{0.6808}{0.0041} \\
fr & 1GB       & \score{0.6710}{0.0060} & \score{0.6823}{0.0034} & \score{0.6773}{0.0041} & \score{0.6899}{0.0040} \\
\bottomrule
\end{tabular}
\caption{XNLI downsampled mean test accuracy with standard deviations across runs.}
\label{tab:xnli-ds-raw-scores-sd}
\end{table*}

\begin{table*}[t]
\centering
\scriptsize
\setlength{\tabcolsep}{5pt}
\renewcommand{\arraystretch}{1.08}
\begin{tabular}{llcccc}
\toprule
& & \multicolumn{2}{c}{Deterministic pretraining} & \multicolumn{2}{c}{PTD pretraining} \\
\cmidrule(lr){3-4}\cmidrule(l){5-6}
Language & PT setup & No FTD & +FTD & No FTD & +FTD \\
\midrule
de & 100MB     & \score{0.7896}{0.0030} & \score{0.7784}{0.0076} & \score{0.7577}{0.0092} & \score{0.7749}{0.0188} \\
de & Bilingual & \score{0.8342}{0.0046} & \score{0.8463}{0.0032} & \score{0.8463}{0.0056} & \score{0.8553}{0.0049} \\
de & 1GB       & \score{0.8386}{0.0052} & \score{0.8451}{0.0048} & \score{0.8556}{0.0062} & \score{0.8641}{0.0044} \\
\addlinespace[2pt]
en & 100MB     & \score{0.8351}{0.0311} & \score{0.8269}{0.0365} & \score{0.8656}{0.0080} & \score{0.8595}{0.0164} \\
en & 1GB       & \score{0.9254}{0.0025} & \score{0.9223}{0.0036} & \score{0.9276}{0.0043} & \score{0.9291}{0.0041} \\
\addlinespace[2pt]
es & 100MB     & \score{0.7734}{0.0055} & \score{0.8099}{0.0095} & \score{0.7988}{0.0061} & \score{0.8093}{0.0080} \\
es & Bilingual & \score{0.8734}{0.0042} & \score{0.8773}{0.0045} & \score{0.8714}{0.0034} & \score{0.8766}{0.0049} \\
es & 1GB       & \score{0.8799}{0.0059} & \score{0.8871}{0.0050} & \score{0.8801}{0.0049} & \score{0.8843}{0.0045} \\
\addlinespace[2pt]
fr & 100MB     & \score{0.7709}{0.0049} & \score{0.7493}{0.0068} & \score{0.7815}{0.0085} & \score{0.8087}{0.0089} \\
fr & Bilingual & \score{0.8783}{0.0049} & \score{0.8768}{0.0033} & \score{0.8706}{0.0035} & \score{0.8786}{0.0032} \\
fr & 1GB       & \score{0.8780}{0.0033} & \score{0.8763}{0.0050} & \score{0.8618}{0.0056} & \score{0.8650}{0.0043} \\
\bottomrule
\end{tabular}
\caption{PAWS-X mean test accuracy with standard deviations across runs.}
\label{tab:pawsx-raw-scores-sd}
\end{table*}
\begin{table*}[t]
\centering
\scriptsize
\setlength{\tabcolsep}{5pt}
\renewcommand{\arraystretch}{1.08}
\begin{tabular}{llcccc}
\toprule
& & \multicolumn{2}{c}{Deterministic pretraining} & \multicolumn{2}{c}{PTD pretraining} \\
\cmidrule(lr){3-4}\cmidrule(l){5-6}
Language & PT setup & No FTD & +FTD & No FTD & +FTD \\
\midrule
de & 100MB     & \score{0.8290}{0.0019} & \score{0.8289}{0.0018} & \score{0.8371}{0.0012} & \score{0.8369}{0.0011} \\
de & Bilingual & \score{0.8515}{0.0016} & \score{0.8514}{0.0015} & \score{0.8521}{0.0015} & \score{0.8520}{0.0017} \\
de & 1GB       & \score{0.8637}{0.0012} & \score{0.8638}{0.0008} & \score{0.8663}{0.0013} & \score{0.8666}{0.0010} \\
\addlinespace[2pt]
en & 100MB     & \score{0.7549}{0.0025} & \score{0.7549}{0.0025} & \score{0.7590}{0.0028} & \score{0.7590}{0.0028} \\
en & 1GB       & \score{0.7917}{0.0015} & \score{0.7916}{0.0015} & \score{0.7954}{0.0019} & \score{0.7953}{0.0018} \\
\addlinespace[2pt]
es & 100MB     & \score{0.8600}{0.0017} & \score{0.8597}{0.0016} & \score{0.8696}{0.0022} & \score{0.8708}{0.0024} \\
es & Bilingual & \score{0.8838}{0.0023} & \score{0.8835}{0.0021} & \score{0.8834}{0.0012} & \score{0.8842}{0.0025} \\
es & 1GB       & \score{0.8907}{0.0019} & \score{0.8910}{0.0019} & \score{0.8918}{0.0018} & \score{0.8912}{0.0015} \\
\addlinespace[2pt]
fr & 100MB     & \score{0.8380}{0.0018} & \score{0.8379}{0.0018} & \score{0.8497}{0.0018} & \score{0.8495}{0.0018} \\
fr & Bilingual & \score{0.8641}{0.0019} & \score{0.8641}{0.0018} & \score{0.8637}{0.0016} & \score{0.8639}{0.0014} \\
fr & 1GB       & \score{0.8716}{0.0020} & \score{0.8718}{0.0020} & \score{0.8720}{0.0017} & \score{0.8718}{0.0019} \\
\addlinespace[2pt]
sw & 100MB     & \score{0.8991}{0.0032} & \score{0.8991}{0.0031} & \score{0.9100}{0.0040} & \score{0.9100}{0.0040} \\
xh & 100MB     & \score{0.8547}{0.0037} & \score{0.8548}{0.0037} & \score{0.8615}{0.0040} & \score{0.8615}{0.0040} \\
\bottomrule
\end{tabular}
\caption{Mean NER F1 scores with standard deviations across runs.}
\label{tab:ner-raw-scores-sd}
\end{table*}

\end{document}